\documentclass[conference]{IEEEtran}
\IEEEoverridecommandlockouts
\usepackage{cite}
\usepackage{amsmath,amssymb,amsfonts}
\usepackage{amsthm}
\usepackage{algorithm}
\usepackage{algpseudocode}
\usepackage{graphicx}
\usepackage{textcomp}
\usepackage{xcolor}
\usepackage{dsfont}
\usepackage{bbold}

\definecolor{graph1}{HTML}{05C090}
\definecolor{graph2}{HTML}{118AB2}

\usepackage[caption=false, font=footnotesize]{subfig}

\usepackage{amsmath} 
\usepackage{tikz}
\usetikzlibrary{arrows,shapes,automata,backgrounds,petri,chains}
\usetikzlibrary{matrix,decorations.pathreplacing, calc, positioning,fit,trees}

\tikzset{>=latex} 
\usepackage{pgfplots} 
\pgfplotsset{compat=1.18}

\newtheorem{definition}{Definition}

\def\BibTeX{{\rm B\kern-.05em{\sc i\kern-.025em b}\kern-.08em
    T\kern-.1667em\lower.7ex\hbox{E}\kern-.125emX}}
\begin{document}

\makeatletter
\newcommand{\linebreakand}{%
  \end{@IEEEauthorhalign}
  \hfill\mbox{}\par
  \mbox{}\hfill\begin{@IEEEauthorhalign}
}
\makeatother

\title{Structural Fusion of Bayesian Networks with Limited Treewidth using Genetic Algorithms\\

}

\author{
    \IEEEauthorblockN{Pablo Torrijos\IEEEauthorrefmark{1}\IEEEauthorrefmark{2}, José A. Gámez\IEEEauthorrefmark{1}\IEEEauthorrefmark{2}, José M. Puerta\IEEEauthorrefmark{1}\IEEEauthorrefmark{2}}
    \IEEEauthorblockA{\IEEEauthorrefmark{1}Instituto de Investigación en Informática de Albacete (I3A). Universidad de Castilla-La Mancha. Albacete, 02071, Spain}
    \IEEEauthorblockA{\IEEEauthorrefmark{2}Departamento de Sistemas Informáticos. Universidad de Castilla-La Mancha. Albacete, 02071, Spain
    \\\{Pablo.Torrijos,Jose.Gamez,Jose.Puerta\}@uclm.es}
}

\maketitle

\begin{center}
\footnotesize
© 2024 IEEE. Personal use of this material is permitted.
Permission from IEEE must be obtained for all other uses.

Published in: 2024 IEEE Congress on Evolutionary Computation (CEC),
pp. 1--8, 2024.
DOI: 10.1109/CEC60901.2024.10611976
\end{center}

\begin{abstract}
This paper introduces an evolutionary computation approach for consensus in structural Bayesian Network (BN) fusion under the constraint of limited treewidth. The consensus BN aims to reconcile multiple input BNs into a single one that retains key structural features present in the original networks. Treewidth, a graph-based parameter associated with computationally tractable inference, is utilized to restrict the complexity of the resulting network. A genetic algorithm is proposed to look for a BN that codifies as much information about the unrestricted fusion as possible while ensuring the treewidth restriction. Experimental evaluation demonstrates the genetic algorithm's ability to obtain consensus BNs with limited treewidth, providing a valuable tool for aggregating information from diverse sources while returning a computationally actionable model.
\end{abstract}

\begin{IEEEkeywords}
Bayesian networks, fusion, consensus, genetic algorithm, combinatorial optimization, federated learning.
\end{IEEEkeywords}

%
%
\section{Introduction} \label{sec:introduction}

A Bayesian Network (BN) \cite{Jensen_Nielsen,Koller_Friedman} is a type of probabilistic graphical model (PGM) that utilizes probability theory to represent uncertainty in specific domains. A BN comprises two key components: a graphical structure encapsulating the interrelationships among domain variables, including their dependencies and independencies, and a collection of parameters in the form of conditional probability tables that quantify the significance of the connections depicted in the graph.
Domain experts can construct BNs by developing both parts (structure and parameters) \cite{Kjaerulff_Madsen}. However, this becomes increasingly challenging as the problem scale expands, often becoming a bottleneck in the knowledge elicitation process. However, learning BNs from data is extensively studied. Several methods have been proposed for this purpose \cite{chickering_optimal_2002,gamez_learning_2011,Scanagatta_review_2019}.

The effectiveness of BNs largely depends on their graphical structure, which is crucial for symbolic (relevance) analysis \cite{relevance_BNs_1997,MeekesRG15} and assists numerical inference processes like belief updating and revision \cite{Jensen_Nielsen}. The combination of their graphical nature and symbolic inference mechanisms categorizes BNs as white-box models, enhancing their interpretability. The growing need for explainable models and the increasing relevance of causal models position BNs at the forefront of technology for addressing knowledge-based problems.

Because many NP-hard problems dealing with BNs can be framed as combinatorial optimization problems, numerous evolutionary computing-based approaches have been proposed to tackle them. This includes learning BNs from data \cite{Larranaga_review_2013}; triangulating the graph to obtain a minimum-sized jointree, thereby optimizing the inference process \cite{GamezPuerta_triangulation_02}; searching for the most probable explanation \cite{CamposGM02}; etc.

This work addresses a different NP-hard problem related to BNs: the structural fusion of BNs \cite{pena_finding_2011}. Given a set of networks defined over the same set of variables, the structural fusion process aims to obtain the consensus model that best represents the input networks \cite{pena_finding_2011, sagrado_qualitative_2003}. This problem is particularly interesting in scenarios where different BNs are generated by different experts or learned from different datasets that cannot be shared, such as in the paradigm of federated learning \cite{Zhang_survey_fl_KBS_2021}. 
Canonical structural fusion produces dense networks, making them practically unusable for reasoning due to the large probability tables created either in the network or in the intermediate steps of inference.

In \cite{Puerta2021Fusion}, a process is described for obtaining the optimal fusion of a set of input networks. In addition, a heuristic is provided (see Section \ref{sec:preliminaries}) that efficiently guides the fusion process and yields results very close to the optimal outcome. The downside of the structural fusion process is that, due to its restrictive definition, it produces dense networks, making them practically unusable for reasoning, as the probability tables created either in the network or in the intermediate steps of inference are very large.

In this work, we propose to relax the restrictions of the canonical definition of BN fusion, bringing it closer to the standard definition of consensus or aggregation in other problems, where each participant gives up something so that the resulting model captures most, but not all, of the restrictions of the input models. Moreover, our goal is to obtain tractable networks for symbolic and numerical reasoning \cite{Benjumeda2015}, so we will restrict the maximum complexity that the consensus-obtained model can have. Thus, the objective of this restricted fusion/consensus will be to obtain the network with a maximum {\em treewidth} (the number of variables in the biggest probability table used at inference) of $tw$ that captures the maximum number of arcs that would be in the model obtained by applying an unrestricted fusion process.

To address the problem, our proposal approach defines structurally constrained fusion as a combinatorial optimization problem and uses a genetic algorithm \cite{Katoch2020} as the search engine. Thus, the main contributions of this work are:
\begin{itemize}
\item Introducing the definition of restricted fusion/consensus of BNs and formulating the associated combinatorial optimization problem.
\item Proposing a greedy algorithm to solve the problem.
\item Designing a genetic algorithm tailored to the BN's restricted fusion problem.
\item Experimentally evaluating the proposal on different benchmarks.
\end{itemize}

The remainder of this paper is organized as follows: Section \ref{sec:preliminaries} describes the problem of BN fusion and discusses its main drawbacks. Next, in Section \ref{sec:problem}, we define the constrained structural fusion of BNs problem, and in Section \ref{sec:proposal}, we describe our evolutionary approach to tackle it. Then, Section \ref{sec:experiments} contains the experimental evaluation carried out, and finally, in Section \ref{sec:conclusions}, we conclude. 

%
%
\section{Preliminaries} \label{sec:preliminaries}

Let ${\cal{X}} = \{X_1, \dots, X_n\}$ be the set of variables in the domain being modeled. A Bayesian network ${\cal{B}} = ({\cal{G}},{\cal{P}})$ is a pair where: ${\cal{G}} = ({\cal{X}},{\cal{A}})$ is a directed acyclic graph (DAG) defined by the set of variables ${\cal{X}}$ as vertices and a set ${\cal{A}}$ of directed arcs between them; and ${\cal{P}}$ is a set of conditional probability distributions, $\{P(X_i|pa(X_i))\}_{i=1}^n$, where $pa(X_i)$ stands for the parent set of $X_i$ in ${\cal{G}}$. Due to the conditional independences represented in the DAG, the joint probability distribution can be factorized as:
\[
P(X_1, \dots, X_n) = \prod_{i=1}^n P(X_i | pa(X_i))
\]

Let $I({\cal{G}})$ represent the set of independences in ${\cal{G}}$ using the {\em d-separation} criterion \cite{Koller_Friedman}. We say that ${\cal{G}}_1$ is an $I$-map of ${\cal{G}}_2$ if $I({\cal{G}}_1) \subseteq I({\cal{G}}_2)$. ${\cal{G}}_1$ is a {\em minimal} $I$-map of ${\cal{G}}_2$ if removing any arc from ${\cal{G}}_1$ it is no longer an $I$-map of ${\cal{G}}_2$.

Given a set of BNs $\{{\cal{B}}_1,\dots,{\cal{B}}_r\}$ and, in particular, their DAGs $\{{\cal{G}}_1, \dots,{\cal{G}}_r\}$, the goal of the structural fusion process \cite{pena_finding_2011} is to obtain a DAG ${\cal{G}}^+$ with the minimal number of arcs such that it is an $I$-map of $\cap_{i=1}^{r} I({\cal{G}}_i)$. In other words, if any conditional independence is codified in ${\cal{G}}^+$, it is also codified in all the input DAGs.

For example, Fig. \ref{fig:redes-originales} shows three DAGs defined over the same four variables, and Fig. \ref{fig:fusion-results} shows two possible results for their fusion. The result on the right is clearly preferred, as it has fewer arcs. In fact, this network represents the optimal fusion for the three input DAGs.

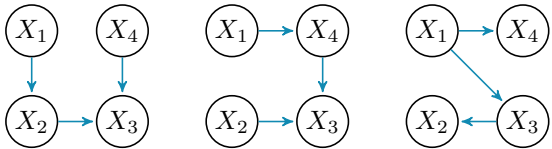
\begin{figure}[b]
    \centering
    \vspace{-0.3cm}
            \begin{tikzpicture}[->,>=stealth',shorten >=1pt,auto,node distance=1.2cm,  
                        semithick]
                  \tikzstyle{every state}=[fill=none,draw=black,text=black]
                
                  \node[state,inner sep=1.5pt,minimum size=1.5pt]         (X1)                      {$X_1$};
                  \node[state,inner sep=1.5pt,minimum size=1.5pt]         (X2) [below of=X1]          {$X_2$};
                  \node[state,inner sep=1.5pt,minimum size=1.5pt]         (X3) [right of=X2]         {$X_3$};
                  \node[state,inner sep=1.5pt,minimum size=1.5pt]         (X4) [above of=X3]          {$X_4$};
                  
                  \path (X1) edge [graph2]             node {} (X2)
                        (X2) edge [graph2]             node {} (X3)
                        (X4) edge [graph2]             node {} (X3);
                \end{tikzpicture}
        \hspace{0.5cm}
            \begin{tikzpicture}[->,>=stealth',shorten >=1pt,auto,node distance=1.2cm,  
                        semithick]
                  \tikzstyle{every state}=[fill=none,draw=black,text=black]
                
                  \node[state,inner sep=1.5pt,minimum size=1.5pt]         (X1)                      {$X_1$};
                  \node[state,inner sep=1.5pt,minimum size=1.5pt]         (X2) [below of=X1]          {$X_2$};
                  \node[state,inner sep=1.5pt,minimum size=1.5pt]         (X3) [right of=X2]         {$X_3$};
                  \node[state,inner sep=1.5pt,minimum size=1.5pt]         (X4) [above of=X3]          {$X_4$};
                  
                  \path (X1) edge [graph2]             node {} (X4)
                        (X2) edge [graph2]             node {} (X3)
                        (X4) edge [graph2]             node {} (X3);
                \end{tikzpicture}
        \hspace{0.5cm}
            \begin{tikzpicture}[->,>=stealth',shorten >=1pt,auto,node distance=1.2cm,  
                        semithick]
                  \tikzstyle{every state}=[fill=none,draw=black,text=black]
                
                  \node[state,inner sep=1.5pt,minimum size=1.5pt]         (X1)                      {$X_1$};
                  \node[state,inner sep=1.5pt,minimum size=1.5pt]         (X2) [below of=X1]          {$X_2$};
                  \node[state,inner sep=1.5pt,minimum size=1.5pt]         (X3) [right of=X2]         {$X_3$};
                  \node[state,inner sep=1.5pt,minimum size=1.5pt]         (X4) [above of=X3]          {$X_4$};
                  
                  \path (X1) edge [graph2]             node {} (X4)
                        (X1) edge [graph2]             node {} (X3)
                        (X3) edge [graph2]             node {} (X2);
                \end{tikzpicture}
    \caption{Input networks for the fusion process.}
    \label{fig:redes-originales}
\end{figure}

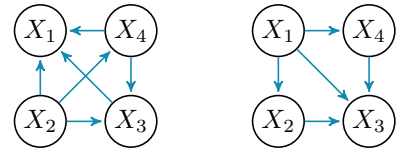
\begin{figure}[b]
    \centering
            \begin{tikzpicture}[->,>=stealth',shorten >=1pt,auto,node distance=1.2cm,  
                        semithick]
                  \tikzstyle{every state}=[fill=none,draw=black,text=black]
                
                  \node[state,inner sep=1.5pt,minimum size=1.5pt]         (X1)                      {$X_1$};
                  \node[state,inner sep=1.5pt,minimum size=1.5pt]         (X2) [below of=X1]          {$X_2$};
                  \node[state,inner sep=1.5pt,minimum size=1.5pt]         (X3) [right of=X2]         {$X_3$};
                  \node[state,inner sep=1.5pt,minimum size=1.5pt]         (X4) [above of=X3]          {$X_4$};
                  
                  \path (X2) edge [graph2]             node {} (X1)
                        (X3) edge [graph2]             node {} (X1)
                        (X4) edge [graph2]             node {} (X1)
                        (X2) edge [graph2]             node {} (X3)
                        (X2) edge [graph2]             node {} (X4)
                        (X4) edge [graph2]             node {} (X3);
                \end{tikzpicture}
        \hspace{1cm}
            \begin{tikzpicture}[->,>=stealth',shorten >=1pt,auto,node distance=1.2cm,  
                        semithick]
                  \tikzstyle{every state}=[fill=none,draw=black,text=black]
                
                  \node[state,inner sep=1.5pt,minimum size=1.5pt]         (X1)                      {$X_1$};
                  \node[state,inner sep=1.5pt,minimum size=1.5pt]         (X2) [below of=X1]          {$X_2$};
                  \node[state,inner sep=1.5pt,minimum size=1.5pt]         (X3) [right of=X2]         {$X_3$};
                  \node[state,inner sep=1.5pt,minimum size=1.5pt]         (X4) [above of=X3]          {$X_4$};
                  
                  \path (X1) edge [graph2]             node {} (X2)
                        (X1) edge [graph2]             node {} (X3)
                        (X1) edge [graph2]             node {} (X4)
                        (X2) edge [graph2]             node {} (X3)
                        (X4) edge [graph2]             node {} (X3);
                \end{tikzpicture}
    \caption{Two different results for the fusion of the three networks in Fig. \ref{fig:redes-originales}.}
    \label{fig:fusion-results}
\end{figure}

Following \cite{pena_finding_2011,sagrado_qualitative_2003}, the fusion process can be carried out as follows:
\begin{enumerate}
\item Let $\sigma$ be an ordering for the variables in ${\cal{X}}$.
\item ${\cal{G}}^{\sigma}_i \leftarrow A({\cal{G}}_i, \sigma)$, for $i=1, \dots, r$.
\item ${\cal{G}}^+ \leftarrow \cup_{i=1}^r arcs({\cal{G}}^{\sigma}_i)$. 
\end{enumerate}
Where $arcs({\cal{G}})$ returns the set of arcs in the received DAG and the $A({\cal{G}}_i, \sigma)$ method \cite{pena_finding_2011} obtains a new DAG compatible with the order $\sigma$ and being a minimal I-map of ${\cal{G}}$.
To do this, the method $A$ needs to reverse some arcs, which usually means adding other arcs (parents) to maintain the maximum set of initial conditional independences. That is, usually ${\cal{G}}^{\sigma}_i$ is a more complex network than ${\cal{G}}_i$. The description of the $A$ method and the theoretical results behind the fusion process are beyond the scope of this article. Hence, the reader is referred to \cite{pena_finding_2011,sagrado_qualitative_2003} for details.

Fig. \ref{fig:redes-transformadas} shows the three DAGs obtained by applying the $A$ method to the three original DAGs in Fig. \ref{fig:redes-originales} using $\sigma = (X_1,X_2,X_4,X_3)$. As can be observed for any node in the three networks, their parents appear early in the ordering $\sigma$. In fact, this is the ordering used to obtain the consensus network shown on the right side of Fig. \ref{fig:fusion-results}.

\begin{figure}[tb]
    \centering
    \vspace{-0.3cm}
            \begin{tikzpicture}[->,>=stealth',shorten >=1pt,auto,node distance=1.2cm,  
                        semithick]
                  \tikzstyle{every state}=[fill=none,draw=black,text=black]
                
                  \node[state,inner sep=1.5pt,minimum size=1.5pt]         (X1)                      {$X_1$};
                  \node[state,inner sep=1.5pt,minimum size=1.5pt]         (X2) [below of=X1]          {$X_2$};
                  \node[state,inner sep=1.5pt,minimum size=1.5pt]         (X3) [right of=X2]         {$X_3$};
                  \node[state,inner sep=1.5pt,minimum size=1.5pt]         (X4) [above of=X3]          {$X_4$};
                  
                  \path (X1) edge [graph2]             node {} (X2)
                        (X2) edge [graph2]             node {} (X3)
                        (X4) edge [graph2]             node {} (X3);
                \end{tikzpicture}
        \hspace{0.5cm}
            \begin{tikzpicture}[->,>=stealth',shorten >=1pt,auto,node distance=1.2cm,  
                        semithick]
                  \tikzstyle{every state}=[fill=none,draw=black,text=black]
                
                  \node[state,inner sep=1.5pt,minimum size=1.5pt]         (X1)                      {$X_1$};
                  \node[state,inner sep=1.5pt,minimum size=1.5pt]         (X2) [below of=X1]          {$X_2$};
                  \node[state,inner sep=1.5pt,minimum size=1.5pt]         (X3) [right of=X2]         {$X_3$};
                  \node[state,inner sep=1.5pt,minimum size=1.5pt]         (X4) [above of=X3]          {$X_4$};
                  
                  \path (X1) edge [graph2]             node {} (X4)
                        (X2) edge [graph2]             node {} (X3)
                        (X4) edge [graph2]             node {} (X3);
                \end{tikzpicture}
        \hspace{0.5cm}
            \begin{tikzpicture}[->,>=stealth',shorten >=1pt,auto,node distance=1.2cm,  
                        semithick]
                  \tikzstyle{every state}=[fill=none,draw=black,text=black]
                
                  \node[state,inner sep=1.5pt,minimum size=1.5pt]         (X1)                      {$X_1$};
                  \node[state,inner sep=1.5pt,minimum size=1.5pt]         (X2) [below of=X1]          {$X_2$};
                  \node[state,inner sep=1.5pt,minimum size=1.5pt]         (X3) [right of=X2]         {$X_3$};
                  \node[state,inner sep=1.5pt,minimum size=1.5pt]         (X4) [above of=X3]          {$X_4$};
                  
                  \path (X1) edge [graph2]             node {} (X2)
                        (X1) edge [graph2]             node {} (X3)
                        (X1) edge [graph2]             node {} (X4)
                        (X2) edge [graph2]             node {} (X3);
                \end{tikzpicture}
    \caption{Networks obtained from those in Fig. \ref{fig:redes-originales} by the $A$ method with $\sigma = (X_1,X_2,X_4,X_3)$.}
    \label{fig:redes-transformadas}
\end{figure}
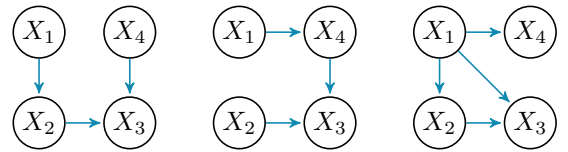

Without delving into the complexity of algorithm $A$ and the theoretical foundations that ensure the fusion process described produces a minimal $I$-map for the input networks, there remains the problem of determining the best possible order to guide the fusion since different orders result in networks of vastly differing quality (see Fig. \ref{fig:fusion-results}). Although this problem is NP-hard, a heuristic has recently been proposed that efficiently yields an order whose use in the fusion process results in networks very close to the optimal \cite{Puerta2021Fusion}. This work assumes that $\sigma$ is calculated using this heuristic.

Despite apparent success in addressing BN fusion practically, the experiments in \cite{Puerta2021Fusion} reveal a notable drawback. The stringent requirement of the BN fusion definition, which operates on an all-or-nothing basis — that is, the resultant network cannot reflect any conditional independence that is not present in any of the networks — leads to the network obtained being so dense and complex that it is often not useful as an actionable model, meaning it cannot be used for symbolic or numerical reasoning. This strict requirement is not present in other consensus-seeking problems, such as group decision-making \cite{Zhang_consensus_2019}, or rank aggregation \cite{Boehmer_rank_aggregation_AAAI_2023}, where the outcome should represent the inputs as best as possible but without the demand to contain strictly all the information provided by them. Based on this more relaxed idea of consensus-seeking and the desire to obtain usable consensus BN models, we define the relaxed BN fusion problem in the next section.

%
%
\section{Problem definition} \label{sec:problem}

Our goal is to achieve a consensus on a model that is as similar as possible to the network resulting from an unrestricted fusion process but usable for reasoning.

To evaluate the first condition, we propose using the distance between two DAGs, ${\cal{G}}_1$ and ${\cal{G}}_2$. The first option could be to use the Hamming distance between two DAGs, measured as the number of edges that would need to be deleted or added to transform ${\cal{G}}_1$ into ${\cal{G}}_2$. Observing the graphs shown in Fig. \ref{fig:fusion-results}, we see that transforming the graph on the left into the one on the right requires removing 4 arcs and adding 3, that is, a distance equal to 7; hence, we can conclude that they are very different graphs. However, in the case of Bayesian Networks (BNs), we can have different graphs that encode almost the same set of independencies according to the d-separation criterion \cite{Jensen_Nielsen}. Therefore, this distance is not suitable for evaluating the distance between BNs. In particular, the two DAGs in Figure \ref{fig:fusion-results} differ only in one conditional independence sentence, as $X_2$ and $X_4$ are conditionally independent given $X_1$ in the simpler DAG but not in the more complex one.
To reflect this fact, the DAG is usually transformed into its {\em moral} graph; that is, undirected edges are added between all the parents of a node, and then the direction of all arcs is removed. This distance is known as Structural Moral Hamming Distance (SMHD) and is commonly used in the literature \cite{Kim2019-moral} to compare BNs.


On the other hand, we will restrict its maximum complexity to ensure that the consensus model obtained is usable in reasoning processes. In this sense, the complexity of inference in BNs is exponential on the number of variables of the largest probability table constructed during the inference process. This number is known as {\em treewidth} and can be obtained from a DAG in a two-step process: (1) obtain the moral graph, and (2) triangulate the moral graph and get the biggest subset of nodes fully connected ({\em clique}) \cite{Koller_Friedman}. The moralization of the graph does not introduce complexity in terms of the size of the biggest clique. However, the triangulation process usually needs to add additional edges, which normally produces bigger cliques than those in the original graph. 

Considering the above criteria and how to measure them, we propose the following definition.

\begin{definition}[Restricted Structural Fusion/Consensus of Bayesian Networks.]\label{def:problem}

Let $\{{\cal{G}}_1, \dots,{\cal{G}}_r\}$ be the DAGs representing the graphical structure of $r$ BNs and ${\cal{G}}^+$ the result of an unrestricted structural fusion of the $r$ Bayesian networks, obtained following the process described in Section \ref{sec:preliminaries}. If $tw \in  \mathds{N}, tw\geq 2$ is the maximum admitted treewidth, the goal of the restricted structural fusion is to obtain
\begin{equation}
{\cal{G}}^+_{tw} = \arg \min_{{\cal{G}}; \,\textrm{treewidth}({\cal{G}}) \leq tw} \textrm{SMHD}({\cal{G}}, {\cal{G}}^+)
\label{eq:problema}
\end{equation}

\end{definition}

Note that all the processes carried out on the input graphs (fusion, moralization, triangulation, etc.) are symbolic processes that do not require computing probability tables so that they can be performed efficiently.

To illustrate the effect of fusion and fusion with limited treewidth, let us consider the example of applying the fusion process to the three graphs shown in Fig. \ref{fig:treewidth}(a-c). We can observe that they are simple graphs with a treewidth of 2. However, the application of unrestricted fusion produces a quite complex network for only 5 variables (Fig. \ref{subfig:BNfusion}), with a treewidth of 5 (a complete graph). On the other hand, if we limit the treewidth to 3, the obtained consensus network is the one shown in \ref{subfig:BNfusionLimited}, which bears a great similarity to that obtained by unrestricted fusion but is much simpler for reasoning. 


\begin{figure}[tb]
    \centering
    \vspace{-0.3cm}
        \subfloat[First BN. \label{subfig:BN1}]{%
            \begin{tikzpicture}[->,>=stealth',shorten >=1pt,auto,node distance=1.1cm,  
                        semithick]
                  \tikzstyle{every state}=[fill=none,draw=black,text=black]
                
                  \node[state,inner sep=1.5pt,minimum size=1.5pt]         (X1)                      {$X_1$};
                  \node[state,inner sep=1.5pt,minimum size=1.5pt]         (X2) [below left of=X1]          {$X_2$};
                  \node[state,inner sep=1.5pt,minimum size=1.5pt]         (X3) [below of=X2]         {$X_3$};
                  \node[state,inner sep=1.5pt,minimum size=1.5pt]         (X4) [below of=X3]          {$X_4$};
                  \node[state,inner sep=1.5pt,minimum size=1.5pt]         (X5) [below right of=X1]          {$X_5$};
                  
                  \path (X1) edge [graph2]             node {} (X2)
                        (X1) edge [graph2]             node {} (X5)
                        (X2) edge [graph2]             node {} (X3)
                        (X3) edge [graph2]             node {} (X4);
                \end{tikzpicture}}
        \hspace{0.2cm}
        \subfloat[Second BN. \label{subfig:BN2}]{%
            \begin{tikzpicture}[->,>=stealth',shorten >=1pt,auto,node distance=1.1cm,
                        semithick]
                  \tikzstyle{every state}=[fill=none,draw=black,text=black]
                
                  \node[state,inner sep=1.5pt,minimum size=1.5pt]         (X1)                      {$X_1$};
                  \node[state,inner sep=1.5pt,minimum size=1.5pt]         (X4) [below left of=X1]          {$X_4$};
                  \node[state,inner sep=1.5pt,minimum size=1.5pt]         (X3) [below right of=X1]         {$X_3$};
                  \node[state,inner sep=1.5pt,minimum size=1.5pt]         (X2) [below of=X4]          {$X_2$};
                  \node[state,inner sep=1.5pt,minimum size=1.5pt]         (X5) [below of=X3]          {$X_5$};
                  
                  \path (X1) edge [graph2]             node {} (X4)
                        (X1) edge [graph2]             node {} (X3)
                        (X4) edge [graph2]             node {} (X2)
                        (X3) edge [graph2]             node {} (X5);
                \end{tikzpicture}}
        \hspace{0.2cm}
        \subfloat[Third BN. \label{subfig:BN3}]{%
            \begin{tikzpicture}[->,>=stealth',shorten >=1pt,auto,node distance=1.1cm,
                        semithick]
                  \tikzstyle{every state}=[fill=none,draw=black,text=black]
                
                  \node[state,inner sep=1.5pt,minimum size=1.5pt]         (X5)                      {$X_5$};
                  \node[state,inner sep=1.5pt,minimum size=1.5pt]         (X3) [below left of=X5]          {$X_3$};
                  \node[state,inner sep=1.5pt,minimum size=1.5pt]         (X1) [below of=X3]         {$X_1$};
                  \node[state,inner sep=1.5pt,minimum size=1.5pt]         (X4) [below of=X1]          {$X_4$};
                  \node[state,inner sep=1.5pt,minimum size=1.5pt]         (X2) [below right of=X5]          {$X_2$};
                  
                  \path (X5) edge [graph2]             node {} (X3)
                        (X5) edge [graph2]             node {} (X2)
                        (X3) edge [graph2]             node {} (X1)
                        (X1) edge [graph2]             node {} (X4);
                \end{tikzpicture}}
        \\
        \vspace{-0.6cm}
        \subfloat[Fusion of the three BNs without treewidth limit (treewidth=5). \label{subfig:BNfusion}]{%
            \begin{tikzpicture}[->,>=stealth',shorten >=1pt,auto,node distance=1.2cm,
                        semithick]
                  \tikzstyle{every state}=[fill=none,draw=black,text=black]
                
                  \node[state,inner sep=1.5pt,minimum size=1.5pt]         (X4)           {$X_4$};
                  \node[state,inner sep=1.5pt,minimum size=1.5pt]         (X2) [below left of=X4]         {$X_2$};
                  \node[state,inner sep=1.5pt,minimum size=1.5pt]         (X3) [below right of=X4]          {$X_3$};
                  \node[state,inner sep=1.5pt,minimum size=1.5pt]         (X5) [below right of=X2]          {$X_5$};
                  \node[state,inner sep=1.5pt,minimum size=1.5pt]         (X1) [above right of=X3]                {$X_1$};
                  
                  \path (X1) edge [graph2, bend right=75]             node {} (X2)
                        (X1) edge [graph2]             node {} (X4)
                        (X1) edge [graph2]             node {} (X3)
                        (X1) edge [graph2, bend left=50]             node {} (X5)

                        (X3) edge [graph2]             node {} (X4)
                        (X3) edge [graph2]             node {} (X5)
                        (X3) edge [graph2]             node {} (X2)

                        (X4) edge [graph2]             node {} (X2)
                        (X5) edge [graph2]             node {} (X2);
                \end{tikzpicture}}
        \hspace{0.1cm}
        \subfloat[Fusion of the three BNs limiting the maximum treewidth to 3. \label{subfig:BNfusionLimited}]{%
            \begin{tikzpicture}[->,>=stealth',shorten >=1pt,auto,node distance=1.2cm,
                        semithick]
                  \tikzstyle{every state}=[fill=none,draw=black,text=black]
                
                  \node[state,inner sep=1.5pt,minimum size=1.5pt]         (X4)           {$X_4$};
                  \node[state,inner sep=1.5pt,minimum size=1.5pt]         (X2) [below left of=X4]         {$X_2$};
                  \node[state,inner sep=1.5pt,minimum size=1.5pt]         (X3) [below right of=X4]          {$X_3$};
                  \node[state,inner sep=1.5pt,minimum size=1.5pt]         (X5) [below right of=X2]          {$X_5$};
                  \node[state,inner sep=1.5pt,minimum size=1.5pt]         (X1) [above right of=X3]                {$X_1$};
                  
                  \path (X1) edge [graph2, bend right=75]             node {} (X2)
                        (X1) edge [graph2]             node {} (X4)
                        (X1) edge [graph2]             node {} (X3)
                        (X1) edge [graph2, bend left=50]             node {} (X5)

                        (X3) edge [graph2]             node {} (X4)
                        (X3) edge [graph2]             node {} (X5)
                        
                        (X5) edge [graph2]             node {} (X2);
                \end{tikzpicture}}
    \caption{Fusion of BNs with treewidth 2 with and without treewidth limit.}
    \label{fig:treewidth}
\end{figure}
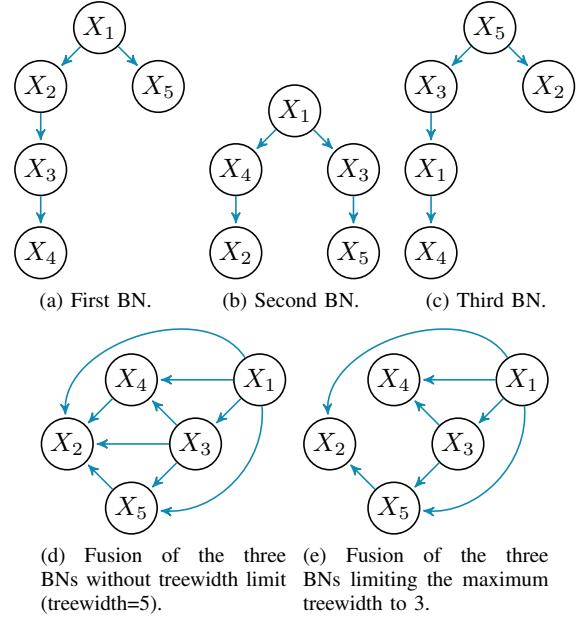


\subsection{Greedy method}\label{ss:greedy}

As an initial strategy to derive a consensus network with constrained treewidth from a collection of input networks, we propose the following greedy method (Algorithm \ref{alg:greedy}), which may serve as a baseline:

\begin{algorithm}
\caption{Greedy BN consensus with limited treewidth}
\label{alg:greedy}
\begin{algorithmic}[1] 
    \Require $\{{\cal{G}}_1, \dots,{\cal{G}}_r\}$ defined over ${\cal{X}}$; $tw \geq 2$
    \Ensure ${\cal{G}}^+_{tw}$
        \State $\sigma \leftarrow$ an ordering for the variables in ${\cal{X}}$ \Comment Use \cite{Puerta2021Fusion}
        \State ${\cal{G}}^{\sigma}_i \leftarrow A({\cal{G}}_i, \sigma)$, for $i=1, \dots, r$
        \State $\textrm{candidates} \leftarrow \cup_{i=1}^r \textrm{arcs}({\cal{G}}^{\sigma}_i)$
        \State $\#(a_{jk}) \leftarrow \sum_{i=1}^r {\mathbb{1}}(a_{jk} \in {\cal{G}}^{\sigma}_i), \; \forall a_{jk} \in \textrm{candidates}$ 
        \State ${\cal{G}}^+_{tw} \leftarrow \textrm{empty graph defined over} \; {\cal{X}}$
        \While { candidates  $\neq \emptyset$ }
            \State Extract $a_{jk}$ with highest $\#(a_{jk})$ from candidates
            \State If $\textrm{treewidth}({\cal{G}}^+_{tw} \cup \{a_{jk}\}) \leq tw$ add $a_{jk}$ to ${\cal{G}}^+_{tw}$
        \EndWhile
        \State Return ${\cal{G}}^+_{tw}$
    \end{algorithmic}
\end{algorithm}

The method iterates through all the candidate arcs, which are ordered by the frequency of their appearance in the transformed DAGs (ties are broken randomly). Consequently, the more frequently an arc appears in the input (transformed) DAGs, the higher the likelihood of it being incorporated into the consensus. 


%
%
\section{Proposal} \label{sec:proposal}

Genetic algorithms and, in general, evolutionary computation \cite{Katoch2020} constitute a very effective tool to deal with machine learning problems \cite{Telikani_review_evolutionary_ML_2021} and other challenging artificial intelligence (AI) problems, as it is eXplainable AI \cite{Bacardit_GA_XAI_2022}. In particular, regarding the BN formalism, several problems have been addressed; among them, we can identify learning them from data \cite{Larranaga_review_2013}, graph triangulation  \cite{GamezPuerta_triangulation_02}, or searching for the most probable explanation \cite{CamposGM02}. On the other hand, evolutionary algorithms have also been used to look for consensus in other problems like rank aggregation \cite{Aledo_EJOR_18} or group decision-making \cite{Benito-consensus-GA_2020}. Encouraged by the success of these previous achievements, in this work, we design a genetic algorithm to deal with the limited fusion/aggregation of BNs.








\subsection{Chromosome representation}\label{ss:representation}

As outlined in the previous section, the fusion process is based on using the transformed DAGs, which are obtained by applying the $A({\cal{G}},\sigma)$ method \cite{pena_finding_2011} to all the incoming original DAGs. We also assume that the result of the unconstrained fusion ${\cal{G}}^+$ is available. Then, we can frame our combinatorial optimization problem as the task of selecting a subset of arcs in ${\cal{G}}^+$ to be included in ${\cal{G}}^+_{tw}$.

Consider ${\cal{A}}^+ = \{ a_{jk} \; | \; a_{jk} \in arcs({\cal{G}}^+)\}$. Let $s$ be the size of ${\cal{A}}^+$ and also note that we arrange the arcs in ${\cal{A}}^+$ using a lexicographical order according to their subscripts; for example, $a_{1,2} \prec a_{1,5} \prec a_{2,1}$. Taking this into account, we select {\em binary} representation for our chromosomes; in particular, a chromosome or individual will be a vector $C$ of size $s$ such that
\[
C[i] = \left\{
\begin{array}{ll}
1 & \textrm{if the i-th arc in} \;{\cal{A}}^+\; \textrm{is included in the graph} \\
0 & \textrm{otherwise} \\
\end{array}
\right.
\]

It should be noted that for a DAG with $n$ nodes, the maximum number of arcs is $\frac{n\cdot(n-1)}{2}$. Clearly, the number of candidate arcs in the fusion will be smaller, but it can still be large enough to create an extensive search space. For instance, the DAG in Fig. \ref{subfig:BNfusion} has 9 arcs, meaning that the candidate arcs (${\cal{A}}^+$) represent 90\% of the possible ones in this case.


\subsection{Fitness function}\label{ss:fitness}

For a given chromosome $C$ we decode it into a DAG ${\cal{G}}_C$ with ${\cal{X}}$ as vertices, and including the i-th arc $a_{jk} \in {\cal{A}}^+$ only if $C[i]=1$. According to Definition \ref{def:problem}, we define the following fitness or evaluation function:
\[
\resizebox{0.5\textwidth}{!}{
$f(C) = \left\{
\begin{array}{ll}
\textrm{SMHD}({\cal{G}}_C,{\cal{G}}^+) & \textrm{if treewidth}({\cal{G}}_C) \leq tw \\
\textrm{SMHD}({\cal{G}}_C,{\cal{G}}^+) \cdot \frac{\textrm{treewidth} ({\cal{G}}_C)}{tw} & \textrm{otherwise} \\
\end{array}
\right.$
}
\]
being our optimization problem to {\em minimize} $f()$. Note that infeasible individuals are severely penalized, discouraging solutions that violate the treewidth constraint and promoting the exploration of solutions that balance structural similarity and adherence to treewidth limits. In any case, invalid individuals are not considered to be the final solution even if they improve on the best value of $f(C)$ so far.


\subsection{Algorithm structure}


The scheme of the proposed genetic algorithm is outlined in Algorithm \ref{alg:genetic}. In the next subsections, we detail any of the components, but in general, we can observe that it follows the steps of a canonical genetic algorithm. The main particularities are that it needs to compute the same information as the greedy method (Algorithm \ref{alg:greedy}), as this information will be used to seed the population, and that elitism is used to retain the best individuals across generations.


\begin{algorithm}
\caption{Genetic BN consensus with limited treewidth}
\label{alg:genetic}
\begin{algorithmic}[1] 
    \Require $\{{\cal{G}}_1, \dots,{\cal{G}}_r\}$ defined over ${\cal{X}}$; $tw \geq 2$; $nIterations$; $popSize$
    \Ensure ${\cal{G}}^+_{tw}$
        \State $\sigma \leftarrow$ an ordering for the variables in ${\cal{X}}$ \Comment Use \cite{Puerta2021Fusion}
        \State ${\cal{G}}^{\sigma}_i \leftarrow A({\cal{G}}_i, \sigma)$, for $i=1, \dots, r$
        \State ${\cal{G}}^+ \leftarrow \cup_{i=1}^r ({\cal{G}}^{\sigma}_i)$ \Comment{unconstrained fusion}
        \State $\textrm{candidates} \leftarrow \textrm{arcs}({\cal{G}}^+)$
        \State $\#(a_{jk}) \leftarrow \sum_{i=1}^r {\mathbb{1}}(a_{jk} \in {\cal{G}}^{\sigma}_i), \; \forall a_{jk} \in \textrm{candidates}$ 
        \State ${C^*, {\cal{G}}_{C^*}} \leftarrow \textrm{null}$ \Comment{best chromosome/DAG seen so far}
        \State ${C^{l}} \leftarrow \textrm{null}$ \Comment{best chromosome in the last population}
        \State $population \leftarrow \textrm{initialization}(popSize, \#(a_{jk}), \textrm{candidates})$
        \For{$i \leftarrow 1 \textrm{ to } nIterations$} 
            \State \textrm{evaluation}($population,{\cal{G}}^+)$ \Comment{Update ${C^*, \cal{G}}_{C^*}, C^l$}
            \State $population \leftarrow \textrm{selection}(population)$
            \State $population \leftarrow \textrm{crossover}(population)$ 
            \State $population \leftarrow \textrm{mutation}(population)$
            \State $population \leftarrow population \cup \{C^*,C^l\}$ \Comment{Elitism}
        \EndFor
        \State Return ${\cal{G}}^*$ as ${\cal{G}}^+_{tw}$
    \end{algorithmic}
\end{algorithm}


\subsection{Initialization} 

In this work, we consider an informed population initialization to seed it with valuable starting points. We use the greedy method defined in Algorithm \ref{alg:greedy} and run it with treewidth $tw$ and $tw-1$. 
The remaining $popSize-2$ individuals are initialized by taking into account the number of times each arc appears in the transformed input DAGs ($\#(a_{jk}$). The following procedure is used: we start with a chromosome representing the empty graph, that is, $C[i]=0, \forall i$. Then, each gene $C[i]$ is set to $1$ with probability $\frac{1}{1-log(x)}$, where $x$ corresponds to the min-max normalization of $\#(a_{jk})$, i.e. the arc codified by the i-th position in $C$ (see Figure \ref{fig:initialization}). Note that the probability would be uniform ($0.5$), i.e., random initialization if all the arcs in candidates have the same frequency $\#(a_{jk})$.

\begin{figure}[tb]
    \centering
        \begin{tikzpicture}[domain=0:1,xscale=6,yscale=1.5]  
          \def\xa{0} \def\xb{1.05}
          \def\ya{0} \def\yb{1.2}
          \def\N{500} 
          \draw[xstep=0.25,ystep=0.25,very thin, color=lightgray]
            (\xa,\ya) grid (1,1);
          \draw[->]
            (\xa,0) -- (\xb,0)
            node[right] {$x$};
          \draw[->]
            (0,\ya) -- (0,\yb)
            node[left] {$y$};
          \draw[] 
            node[below,scale=0.9] at ( 0,     0) {$0$}
            node[below,scale=0.9] at ( 0.25,   0) {$0.25$}
            node[below,scale=0.9] at ( 0.5,     0) {$0.5$}
            node[below,scale=0.9] at ( 0.75,   0) {$0.75$}
            node[below,scale=0.9] at ( 1,     0) {$1$};
          \draw[] 
            node[left,scale=0.9] at ( 0,  0.25) {$0.25$}
            node[left,scale=0.9] at ( 0,  0.5) {$0.5$}
            node[left,scale=0.9] at ( 0,  0.75) {$0.75$}
            node[left,scale=0.9] at ( 0,  1) {$1$};
          \draw[color=graph2,thick,samples=\N,domain=0.001:1]
            plot(\x,{1/(1-(log10(\x)))})
            node[above right] at (0.48,0.14) {$\displaystyle \mathbf{\frac{1}{1-\log x}}$};
        \end{tikzpicture}
    \caption{Probability of adding a candidate arc $\#(a_{jk})$ in the population informed initialization. $x$ stands for the normalized $\#(a_{jk})$ value.}
    \label{fig:initialization}
\end{figure}
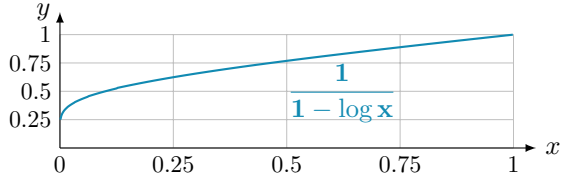

\subsection{Selection}

This work considers tournament selection of size $2$, which is known to maintain diversity. The following process is repeated $popSize$ times: two chromosomes are randomly selected from the current population, and the winner, according to the individual's fitness, is copied to the selected population.


\subsection{Crossover}

Chromosomes are paired by following the order in which they have been selected. Then, a one-point standard crossover for binary representation is applied with probability 1.0, i.e., all the pairs are actually combined.


\subsection{Mutation}

In our proposal, all the chromosomes go through the mutation operation. However, we carefully set the mutation probability of adding/removing an arc to influence the exploration of the search space and the discovery of diverse and effective solutions across generations. We base arc addition/deletion probabilities on the balance between the treewidth of the individual under consideration and the maximum allowed treewidth. Intuitively, the idea is to foster arc addition (deletion) when the individual's treewidth is smaller (greater) than the maximum allowed treewidth. Thus, we start by computing $x$ as the ratio $\textrm{treewidth}({\cal{G}}_C) / tw$, and use this value to compute the mutation probabilities for adding/deleting an arc as shown in Figures \ref{subfig:mutationAdd} and \ref{subfig:mutationRemove}, respectively. 


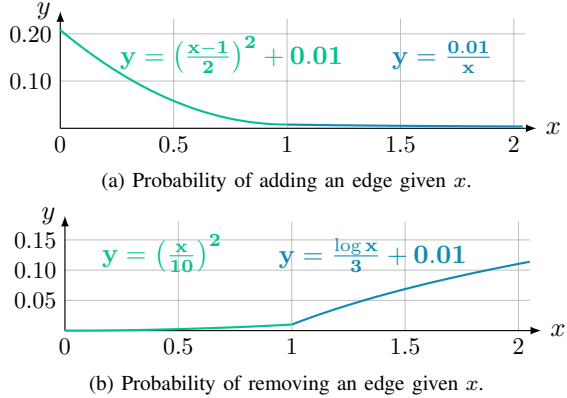
\begin{figure}[tb]
    \centering
        \subfloat[Probability of adding an edge given $x$. \label{subfig:mutationAdd}]{%
            \begin{tikzpicture}[domain=0:2,xscale=3,yscale=5]  
              \def\xa{0} \def\xb{2.1}
              \def\ya{0} \def\yb{0.31}
              \def\N{100} 
              \draw[xstep=0.5,ystep=0.125,very thin, color=lightgray]
                (\xa,\ya) grid (2.05,0.29);
              \draw[->]
                (\xa,0) -- (\xb,0)
                node[right] {$x$};
              \draw[->]
                (0,\ya) -- (0,\yb)
                node[left] {$y$};
              \draw[] 
                node[below,scale=0.9] at ( 0,     0) {$0$}
              	node[below,scale=0.9] at ( 0.5,   0) {$0.5$}
              	node[below,scale=0.9] at ( 1,     0) {$1$}
              	node[below,scale=0.9] at ( 1.5,   0) {$1.5$}
              	node[below,scale=0.9] at ( 2,     0) {$2$};
              \draw[] 
                node[left,scale=0.9] at ( 0,  0.25) {$0.20$}
                node[left,scale=0.9] at ( 0,  0.125) {$0.10$};
              \draw[color=graph1,thick,samples=\N,domain=0:1]
                plot(\x,{((\x - 1)/2)^2 + 0.01})
                node[above right] at (0.22,0.12) {$\mathbf{y = \left(\frac{x - 1}{2}\right)^2 + 0.01}$};
              \draw[color=graph2,thick,samples=\N,domain=1:2.04]
                plot(\x,{0.01/(\x)})
                node[above right] at (1.42,0.12) {$\mathbf{y = \frac{0.01}{x}}$};
            \end{tikzpicture}
            }
        \\
        \vspace{-0.3cm}
        \subfloat[Probability of removing an edge given $x$. \label{subfig:mutationRemove}]{%
            \begin{tikzpicture}[domain=0:2,xscale=3,yscale=8]  
              \def\xa{0} \def\xb{2.1}
              \def\ya{0} \def\yb{0.19}
              \def\N{100} 
              \draw[xstep=0.5,,ystep=0.05,very thin, color=lightgray]
                (\xa,\ya) grid (2.05,0.18);
              \draw[->]
                (\xa,0) -- (\xb,0)
                node[right] {$x$};
              \draw[->]
                (0,\ya) -- (0,\yb)
                node[left] {$y$};
              \draw[] 
                node[below,scale=0.9] at ( 0,     0) {$0$}
              	node[below,scale=0.9] at ( 0.5,   0) {$0.5$}
              	node[below,scale=0.9] at ( 1,     0) {$1$}
              	node[below,scale=0.9] at ( 1.5,   0) {$1.5$}
              	node[below,scale=0.9] at ( 2,     0) {$2$};
              \draw[] 
                node[left,scale=0.9] at ( 0,  0.15) {$0.15$}
                node[left,scale=0.9] at ( 0,  0.1) {$0.10$}
                node[left,scale=0.9] at ( 0,  0.05) {$0.05$};
              \draw[color=graph1,thick,samples=\N,domain=0:1]
                plot(\x,{((\x /10)^2)})
                node[above right] at (0.12,0.081) {$\mathbf{y = \left(\frac{x}{10}\right)^2}$};
              \draw[color=graph2,thick,samples=\N,domain=1:2.05]
                plot(\x,{(log10(\x )/3)+0.01})
                node[above right] at (0.9,0.081) {$\mathbf{y =\frac{\log x}{3}+0.01}$};
            \end{tikzpicture}}
    \caption{Mutation probabilities $(y)$ given $x = \textrm{treewidth}(\cal{G_C}) /$$tw$.}
    \label{fig:mutation}
\end{figure}


\subsection{Population updating}

Following selection, crossover, and mutation, we ensure genetic diversity and retain promising solutions by replacing the two least fit individuals with the best individual from the entire evolutionary process $(C^*)$ and the best individual from the previous generation $(C^l)$. This elitist selection strategy ensures that genetic material with desirable traits persists across generations.

%
%
\section{Experimental evaluation} \label{sec:experiments}
This section outlines the experiments conducted to evaluate our proposed methods.

\subsection{Methodology} \label{subsec:methodology}
We employ a methodology aligned with recent research \cite{Puerta2021Fusion} to conduct our experiments. The goal is to evaluate the proposed methods utilizing synthetic (Experiment 1) and real-world (Experiment 2) networks. 

Ensuring similarity among the input DAGs is crucial, indicating that they encode similar sets of conditional independence statements. 
Thus, we start with a base network (${\cal{G}}_0$), randomly generated in the synthetic case (following \cite{Melanon2004}) and downloaded from \texttt{bnlearn}'s Bayesian Network Repository\footnote{https://www.bnlearn.com/bnrepository/} in the real-world case. Then, we slightly modify ${\cal{G}}_0$ by applying a number of perturbations ($p$) to obtain the input DAGs $\{{\cal{G}}_1, \ldots, {\cal{G}}_r\}$ for the fusion process.
In each perturbation, a pair of nodes $\{X_i,X_j\}$ is randomly selected, and the arc $X_i\rightarrow X_j$ is added or deleted depending on its existence or not in the current DAG. Care is taken to avoid introducing directed cycles. 

Several scenarios are considered for DAG generation. In Experiment 1, we get 9 configurations:  $n = \{10,25,50\}$ nodes and $r = \{10,20,30\}$ input DAGs; while in Experiment 2, we only consider 3 configurations, as the number of nodes is given by the real network considered. Then, the input DAGs for the fusion process are generated by applying $p = n \cdot 0.75$ perturbations over ${\cal{G}}_0$. In any case, as in \cite{Puerta2021Fusion}, we ensure that the maximum number of parents (children) per node is 3 (4) and that the number of arcs in the resulting DAG is limited by $e = n \cdot 2.5$. Finally, we generated 10 sets of DAGs for each configuration to minimize result variability, i.e. the GA is run 10 times for each configuration.

As reported statistics, we compare the SMHD score of the greedy and genetic algorithms, both constrained to given treewidth, against the unconstrained fusion ${\cal{G}}^+$ result. To assess its impact, we evaluate the genetic algorithm through $1000$ iterations with varying population sizes ($popSize = {10,20,50,100}$). We systematically vary treewidth limits ($tw$) from 2 to treewidth(${\cal{G}}^+) - 1$ for each DAG set, excluding $tw=1$ (empty graph) and $tw$ matching the treewidth of the complete fusion. This thorough analysis offers insights into algorithmic performance across different treewidth constraints.

\begin{figure*}[tb]
    \centerline{\includegraphics[width=0.9\linewidth]{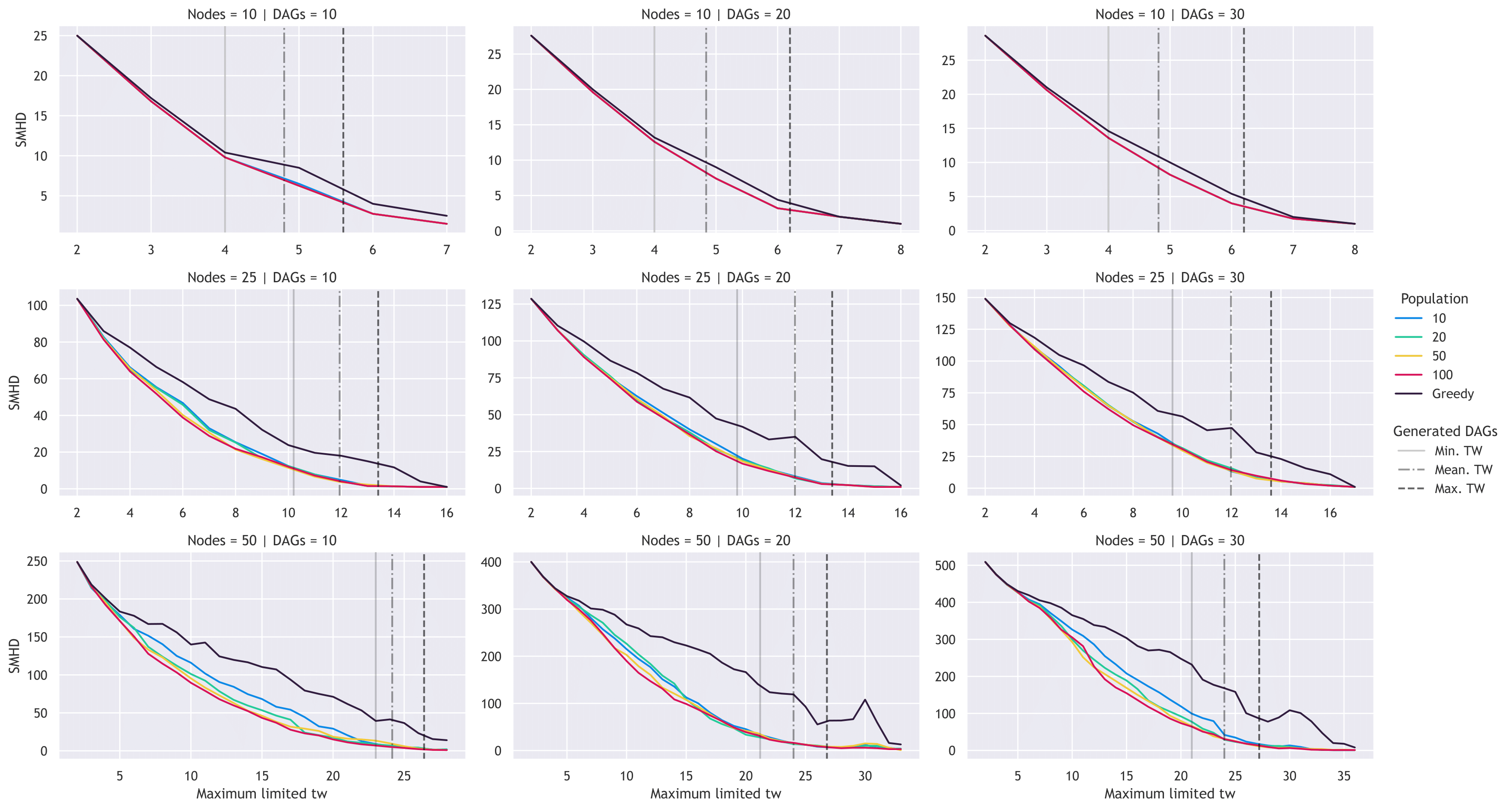}}
    \caption{SMHD obtained by the genetic and the greedy algorithms in the sampled BNs.}
    \label{fig:SMHD}
\end{figure*}

\subsection{Reproducibility} \label{subsec:reproducibility}

All the code presented in this paper is implemented in Java (OpenJDK 17) using the Tetrad 7.1.2-2 causal reasoning library\footnote{https://github.com/cmu-phil/tetrad/releases/tag/v7.1.2-2}. For the convenience of replication, the complete code is made available on GitHub\footnote{https://github.com/ptorrijos99/GeneticTWFusionBN}.

\subsection{Experiment 1 results} \label{subsec:ex1results}

\begin{figure*}[tb]
    \centerline{\includegraphics[width=0.9\linewidth]{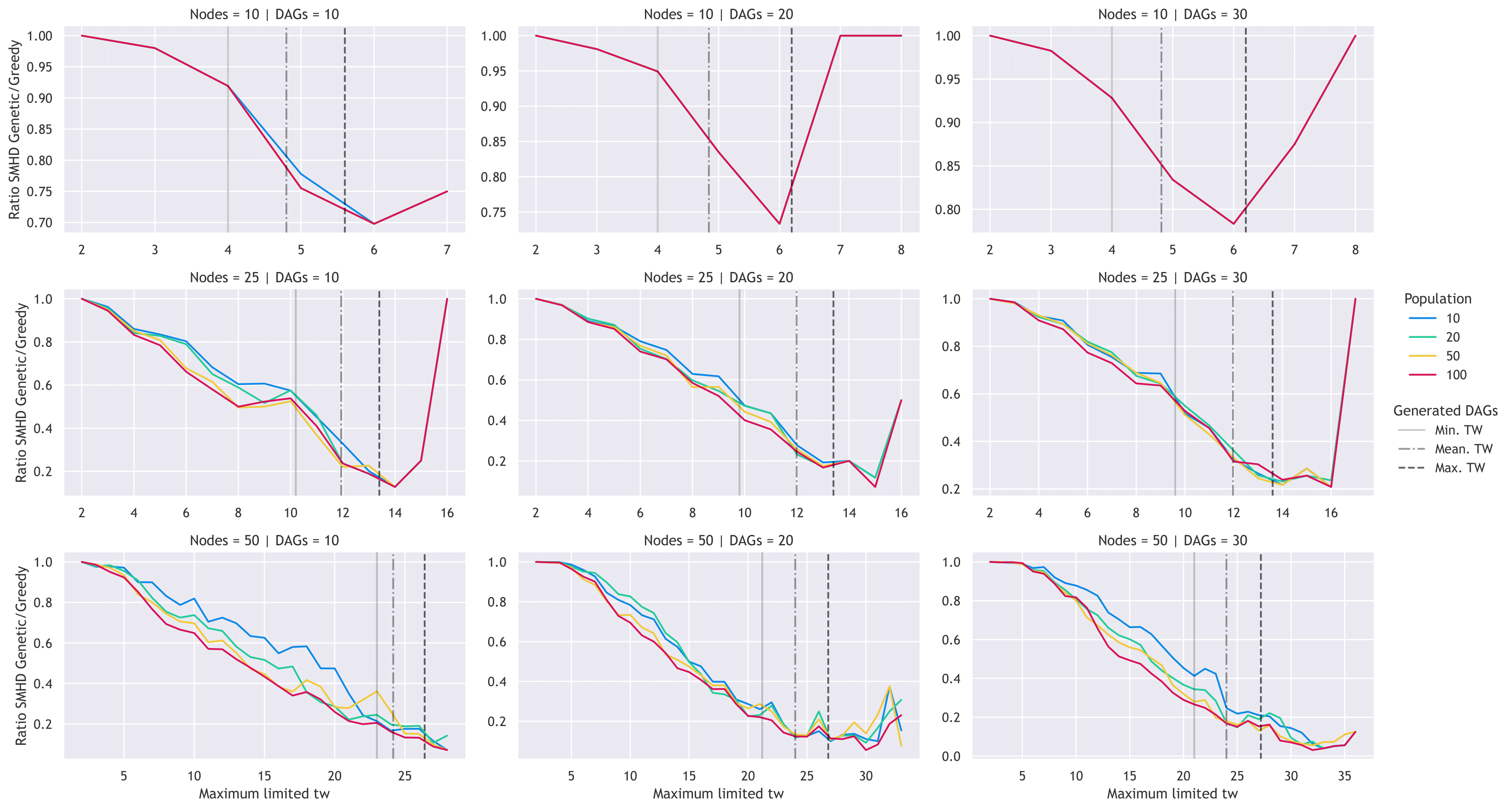}}
    \caption{Ratio between the SMHD obtained by the genetic and the greedy algorithms in the sampled BNs.}
    \label{fig:ratioSMHD}
\end{figure*}

Fig. \ref{fig:SMHD} illustrates the SMHD evolution between the different BNs generated by the greedy and the genetic algorithm (with population sizes 10, 20, 50, and 100) compared to the unconstrained fusion ${\cal{G}}^+$. Treewidth is restricted from 2 to treewidth$({\cal{G}}^+) - 1$. Meanwhile, Fig. \ref{fig:ratioSMHD} illustrates the ratio of the genetic algorithm to the greedy one in terms of the SMHD for the resultant fusion ${\cal{G}}^+_{tw}$ with respect to ${\cal{G}}^+$.

Each picture also depicts the minimum, average, and maximum treewidth of the generated DAGs. These statistics represent the average values over the 10 sets of generated DAGs, offering insights into the treewidth evolution concerning the original networks.

From the analysis of Fig. \ref{fig:SMHD} and Fig. \ref{fig:ratioSMHD}, several conclusions can be drawn:
\begin{itemize}
    \item The genetic algorithm consistently outperforms the greedy approach, with the extent of the improvement being more pronounced when there is a greater margin for enhancement (higher SMHD between ${\cal{G}}^+$ and ${\cal{G}}^*_{tw}$). For 10-node BNs, the genetic algorithm exhibits marginal improvement over the greedy one, with similar performance across all population sizes. This aligns with the limited search space for very low $tw$ values (e.g., 2 or 3) or for high values of $tw$ (e.g., 6, 7, or 8) that are too close to ${\cal{G}}^+$ actual treewidth. In contrast, for 25 and 50 nodes, where the treewidth of ${\cal{G}}^+$ is significantly larger, the genetic algorithm consistently demonstrates substantial improvement over the greedy one. This fact can be seen even more clearly in Fig. \ref{fig:ratioSMHD}, where the improvement is notably bigger for high $tw$ values, reaching exceptional ratios like 0.2 (i.e. five times better). On the other hand, the ratios are close to 1 when $tw$ is too small or too close to ${\cal{G}}^+$ treewidth, that is, when the search space is too limited, so both methods behave similarly.
    
    

    \item Regarding population sizes, as expected, larger populations allow more individual diversity and yield better results, particularly in the more complex scenarios. 
    

    \item It is crucial to highlight the high treewidth of randomly generated DAGs, despite stringent restrictions imposed on the maximum number of parents and children per node. This is evident in the case of networks with 50 nodes, where the average minimum treewidth of the generated networks remains notably high, staying above 20. Merging these high-treewidth DAGs in ${\cal{G}}^+$ results in a network with an even higher treewidth. Consequently, constrained structural fusion becomes more relevant to obtaining tractable BNs. The genetic algorithm achieves BNs structurally almost identical to ${\cal{G}}^+$ while maintaining a treewidth similar to the average treewidth of the randomly generated DAGs. Moreover, this is where the greatest improvement over the greedy algorithm occurs.

\end{itemize}

\subsection{Experiment 2 results} \label{subsec:ex2results}

In this experiment, we consider two manageable real-world networks: {\em Child} (20 nodes, 25 arcs, 230 parameters/probabilities, and max parents equals to 2) and {\em Insurance} (27 nodes, 52 arcs, 1008 parameters/probabilities, and max parents equals to 3).

The sets of DAGs used as input for the fusion process are generated as for the synthetic networks case, but taking the actual (downloaded) DAG as ${\cal{G}}_0$. Fig. \ref{fig:maxParentsreal} shows the increasing number of parents (max number) in the obtained ${\cal{G}}^+_{tw}$ network as a function of the maximum allowed treewidth ($tw$). As can be observed, the complexity of the network quickly increases, even considering networks with a small number of max parents as the original one. Finally, Fig. \ref{fig:SMHDreal} shows the SHMD average value obtained for the greedy and genetic algorithms when applied to the 10 sets of generated networks. As we can see, the same conclusions drawn for the synthetic networks also apply in this case. 


\begin{figure*}[tb]
    \centerline{\includegraphics[width=0.9\linewidth]{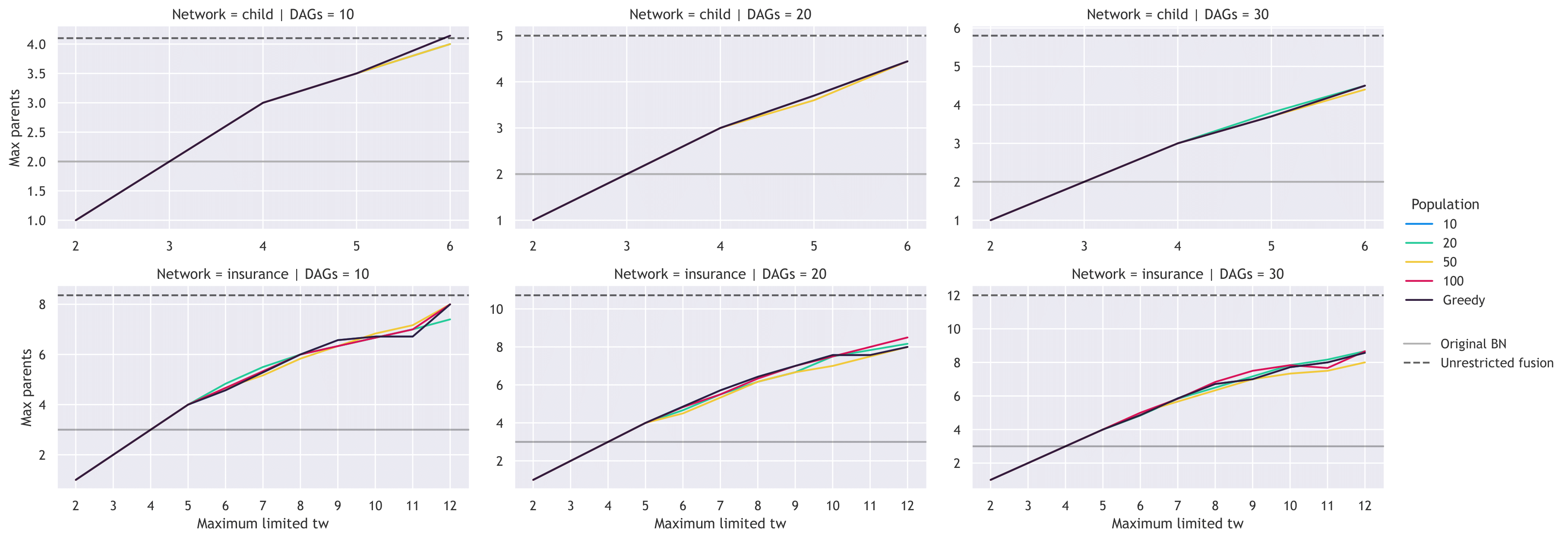}}
    \caption{Mean maximum number of parents in the networks obtained (${\cal{G}}^+_{tw}$) for {\em Child} and {\em Insurance} BNs.}
    \label{fig:maxParentsreal}
\end{figure*}

\begin{figure*}[tb]
    \centerline{\includegraphics[width=0.9\linewidth]{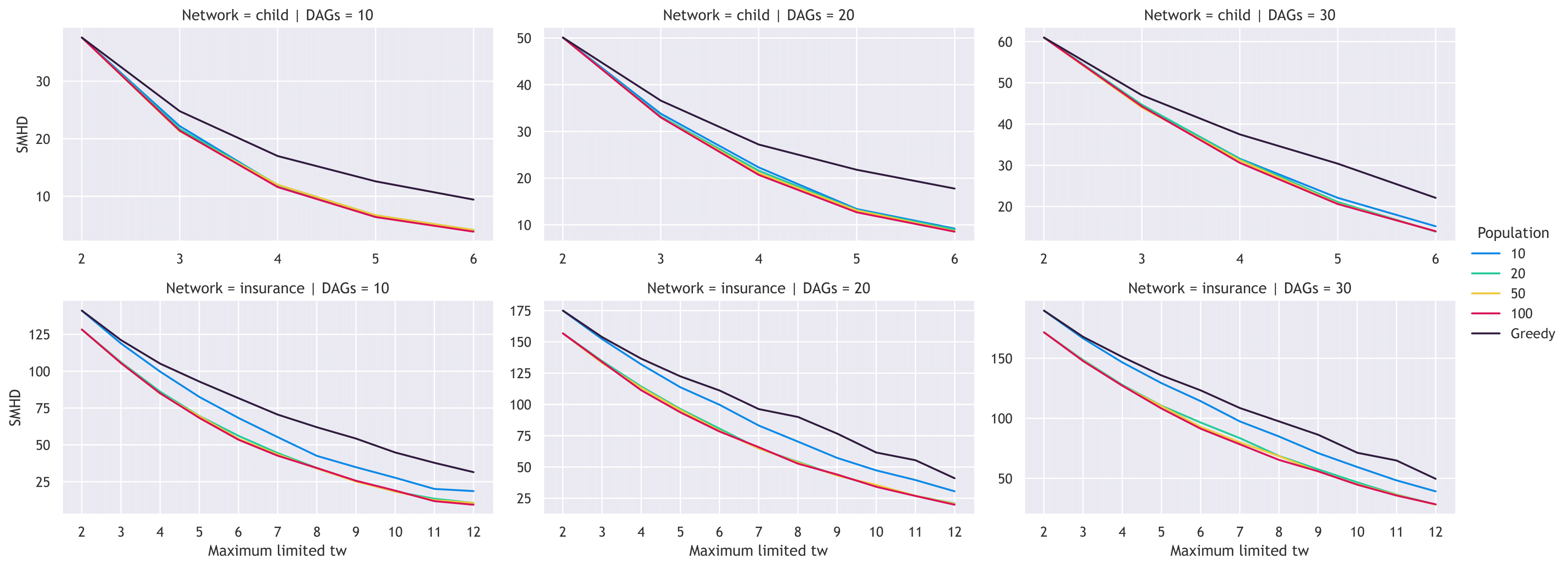}}
    \caption{SMHD obtained by the genetic and the greedy algorithms for {\em Child} and {\em Insurance} BNs.}
    \label{fig:SMHDreal}
\end{figure*}

%
%
\section{Conclusions} \label{sec:conclusions}

In this study, we introduced a genetic algorithm designed for the constrained structural fusion of BNs according to a limited treewidth. The algorithm emerged as a robust solution to the challenges posed by complex networks with high treewidth resulting from multiple DAGs fusion.

Examining randomly generated DAGs underscores the challenge of handling networks with significant treewidth, despite stringent constraints on parent-child relationships. This underscores the need for constrained fusion methods to derive tractable BNs. The genetic algorithm showcased its capability to generate networks resembling unconstrained fusion while preserving treewidths akin to those in the input DAGs. Its consistent outperformance of the baseline greedy algorithm further strengthens its effectiveness.

In conclusion, our proposed genetic algorithm stands out as a promising avenue for constrained BN fusion. It presents a valuable tool for applications where the resulting BN is intended for symbolic or numerical reasoning—scenarios where full fusion proves impractical. This work opens the door for future research on limited/approximate BN aggregation/consensus problems and methods.

\section*{Acknowledgements}
The following projects have funded this work: SBPLY/21/180225/000062 (Government of Castilla-La Mancha and ERDF funds); PID2019--106758GB--C33, TED2021-131291B-I00, FPU21/01074 and PID2022-139293NB-C32 (MCIN/AEI/10.13039/501100011033 and ERDF Next Generation EU); 2022-GRIN-34437 (Universidad de Castilla-La Mancha and ERDF funds).


\bibliographystyle{ieeetr}
\bibliography{biblio}

\end{document}